\documentclass[letterpaper, 10 pt, conference]{ieeeconf}  % Comment this line out if you need a4paper

\IEEEoverridecommandlockouts                              % This command is only needed if 
\usepackage{amsmath} % assumes amsmath package installed
\usepackage{graphics} % for pdf, bitmapped graphics files
\usepackage{epsfig} % for postscript graphics files
\usepackage{times} % assumes new font selection scheme installed
\usepackage{amsmath} % assumes amsmath package installed
\usepackage{amssymb}  % assumes amsmath package installed
\usepackage{hyperref} % for url links
\usepackage{xcolor} % for text in colors
\usepackage{arydshln} % dashlines in tables
\usepackage{color} 
\usepackage{booktabs}
\usepackage{hyperref}
\usepackage{float}
\usepackage{multirow}
\usepackage{pifont}
\usepackage[skip=0pt,font=footnotesize]{caption}
\usepackage{subcaption}
\usepackage{bm}
\usepackage{tikz}
\usepackage{tikz-3dplot}
\newcommand{\iros}[1]{{\color{black} #1}}

\newcommand{\wfr}[0]{\ensuremath{W}} % world frame
\newcommand{\bfr}[0]{\ensuremath{B}} % body frame
\newcommand{\rotateRPY}[3]% roll, pitch, yaw
{   \pgfmathsetmacro{\rollangle}{#1}
    \pgfmathsetmacro{\pitchangle}{#2}
    \pgfmathsetmacro{\yawangle}{#3}

    \pgfmathsetmacro{\newxx}{cos(\yawangle)*cos(\pitchangle)}
    \pgfmathsetmacro{\newxy}{sin(\yawangle)*cos(\pitchangle)}
    \pgfmathsetmacro{\newxz}{-sin(\pitchangle)}
    \path (\newxx,\newxy,\newxz);
    \pgfgetlastxy{\nxx}{\nxy};

    \pgfmathsetmacro{\newyx}{cos(\yawangle)*sin(\pitchangle)*sin(\rollangle)-sin(\yawangle)*cos(\rollangle)}
    \pgfmathsetmacro{\newyy}{sin(\yawangle)*sin(\pitchangle)*sin(\rollangle)+ cos(\yawangle)*cos(\rollangle)}
    \pgfmathsetmacro{\newyz}{cos(\pitchangle)*sin(\rollangle)}
    \path (\newyx,\newyy,\newyz);
    \pgfgetlastxy{\nyx}{\nyy};

    \pgfmathsetmacro{\newzx}{cos(\yawangle)*sin(\pitchangle)*cos(\rollangle)+ sin(\yawangle)*sin(\rollangle)}
    \pgfmathsetmacro{\newzy}{sin(\yawangle)*sin(\pitchangle)*cos(\rollangle)-cos(\yawangle)*sin(\rollangle)}
    \pgfmathsetmacro{\newzz}{cos(\pitchangle)*cos(\rollangle)}
    \path (\newzx,\newzy,\newzz);
    \pgfgetlastxy{\nzx}{\nzy};
}
\DeclareMathOperator*{\argmin}{arg\,min}

\definecolor{somegray}{rgb}{0.5, 0.5, 0.5}
\newcommand{\darkgrayed}[1]{\textcolor{somegray}{#1}}
\makeatletter
\newcommand*\titleheader[1]{\gdef\@titleheader{#1}}
\AtBeginDocument{%
  \let\st@red@title\@title
  \def\@title{%
    \vskip-3em
    \bgroup\normalfont\large\centering\@titleheader\par\egroup
    \vskip1.5em\st@red@title}
}
\makeatother

\titleheader{\darkgrayed{This paper has been accepted for publication at the\\
IEEE/RSJ International Conference on Intelligent Robots (IROS), Detroit, 2023.
\copyright IEEE}}

\title{\LARGE \bf
Autonomous Power Line Inspection with Drones\\
via Perception-Aware MPC
}

\author{Jiaxu Xing$^{*}$, Giovanni Cioffi$^{*}$, Javier Hidalgo-Carri\'o, and Davide Scaramuzza% <-this % stops a space
    \thanks{$^{*}$ These authors contributed equally.
    The authors are with the Robotics and Perception Group, Department of Informatics, University of Zurich, and Department of Neuroinformatics, University of Zurich and ETH Zurich, Switzerland (\protect\url{http://rpg.ifi.uzh.ch}). 
    This work was supported by the Swiss National Science Foundation (SNSF) through the National Centre of Competence in Research (NCCR) Robotics, the European Union’s Horizon 2020 Research and Innovation Programme under grant agreement No. 871479 (AERIAL-CORE), and the European Research Council (ERC) under grant agreement No. 864042 (AGILEFLIGHT).
    }% <-this % stops a space
}

\begin{document}
\makeatletter
\g@addto@macro\@maketitle{
  \captionsetup{type=figure}\setcounter{figure}{0}
  \def\mycolspace{1.2mm}
  \centering
    \includegraphics[trim=0.0 60.0 60.0 100.0, clip, width=2.0\columnwidth]{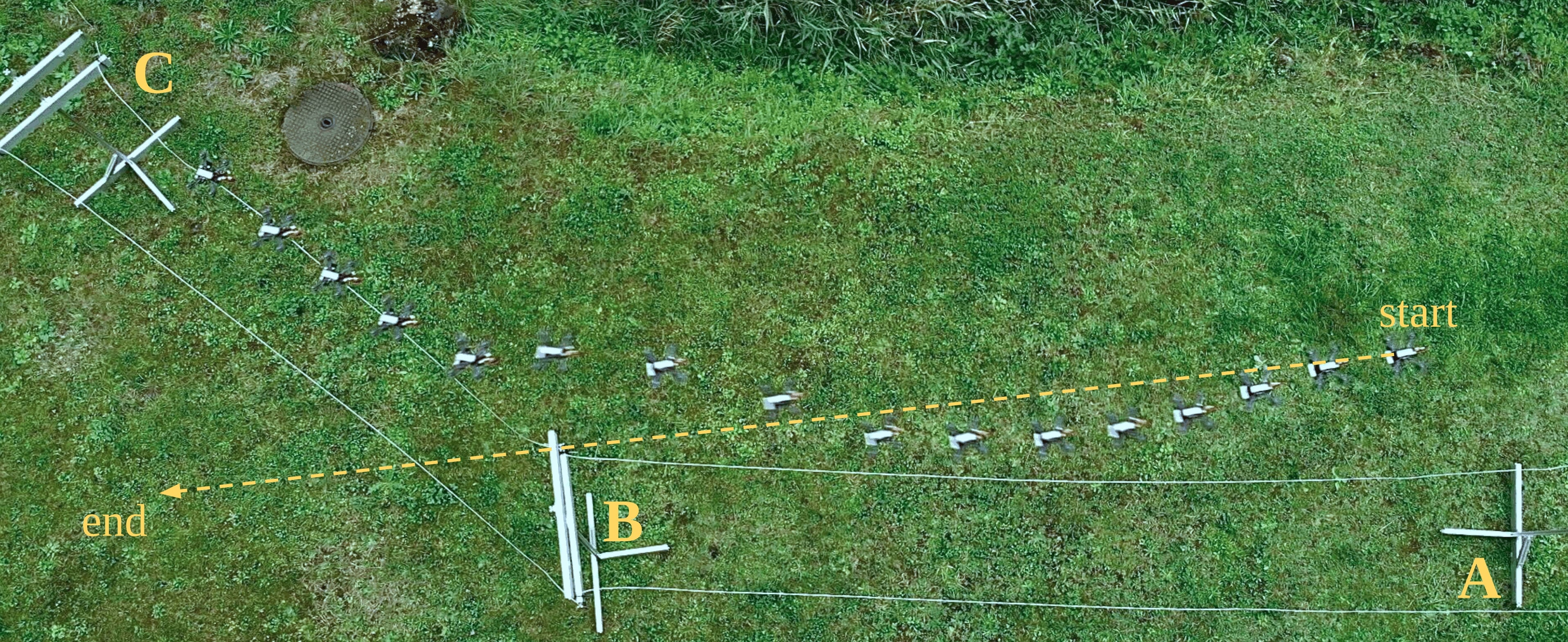}
    \vspace{1.0ex}
	\captionof{figure}{A quadrotor performing power line inspection in a power line test environment with three masts (labelled: A, B, and C) using our 
	proposed approach. The original rough reference trajectory is depicted in yellow and the drone
    deviates from it while avoiding obstacles (the power masts) and keeping the power line visible in the field
    of view of the onboard camera.
	\label{fig:catcheye}}
}
\makeatother
\maketitle
\thispagestyle{empty}
\pagestyle{empty}
%%%%%%%%%%%%%%%%%%%%%%%%%%%%%%%%%%%%%%%%%%%%%%%%%%%%%%%%%%%%%%%%%%%%%%%%%%%%%%%%
\begin{abstract}

Drones have the potential to revolutionize power line inspection by increasing productivity, reducing inspection time, improving data quality, and eliminating the risks for human operators.
Current state-of-the-art systems for power line inspection have two shortcomings:
(i) control is decoupled from perception and needs accurate information about the location of the power lines and masts;
(ii) obstacle avoidance is decoupled from the power line tracking, which results in poor tracking in the vicinity of the power masts, and, consequently, in decreased data quality for visual inspection. 
In this work, we propose a model predictive controller (MPC) that overcomes these limitations by tightly coupling perception and action.
Our controller generates commands that maximize the visibility of the power lines while, at the same time, safely avoiding the power masts.
For power line detection, we propose a lightweight learning-based detector that is trained only on synthetic data and is able to transfer zero-shot to real-world power line images.
We validate our system in simulation and real-world experiments on a mock-up power line infrastructure.
We release our code and datasets to the public.
\end{abstract}
%-------------------------------------------------------------------------
% SECTIONS
\section*{Supplementary Material}\label{sec:SupplementaryMaterial}
\textbf{Video: }\href{https://youtu.be/JA6h-Nv29pU}{https://youtu.be/JA6h-Nv29pU}

\textbf{Code: }\href{https://github.com/uzh-rpg/pampc_for_power_line}{https://github.com/uzh-rpg/pampc\_for\_power\_line}

\section{Introduction} \label{sec:Introduction}
%

%
% How to write abstract and introduction, answer the questions:

% \begin{itemize}
    % \item What is the problem?
    % \item Why is it important?
    % \item Why is the problem hard? What makes it challenging?
    % \item How far has existing work come? What is the next frontier?
    % \item Why hasn’t the problem been solved? What is the stumbling block?
    % \item What does our paper contribute?
    % \item What is the key idea? What is the magic trick? What is the new insight or technique that enables us to advance the frontier?
    % \item What do the experiments say?
% \end{itemize}

%What is the problem?
Drones exhibit the potential to bring about a revolutionary transformation in the industrial inspection market~\cite{DroneiiWP18, eu}. 
Particularly, quadrotors are a fast-to-deploy and cost-effective solution for power line inspection. 
The EU and US power systems consist of more than 10 million km of power lines and distribution transformers, which connect more than 400 million customers~\cite{USEnergy, EUEnergy}.
The power line infrastructure needs to be inspected regularly to avoid power outages and natural disasters (California's second-largest wildfire was sparked when power lines came in contact with a tree~\cite{Wildfire}).

Power line inspection requires teams of specialized human labor who use ad-hoc equipment, such as ropes and scaffoldings, to access the power line infrastructure and manned helicopters for long-range operations. 
These factors result in high expenses, significant dangers for human operators, and low productivity in the inspection operation. 
%Why is it important?
Quadrotors can cut the costs by $50\%$~\cite{DroneiiWP18} as a consequence of the increase of productivity, the reduction of the inspection time, the improvement of the data quality, and the elimination of the risks for the human operators~\cite{DanishUAS2016}. 

%Why is the problem hard? What makes it challenging?
Power line inspection with drones is still done by skilled pilots.
The use of autonomous drones is still limited. 
Developing a fully autonomous solution is challenging because the drone needs to take decisions during the flight that are affected by conflicting goals: the drone needs to fly close to the power lines to maximize the data quality for visual inspection starting from a rough knowledge of the reference trajectory, and at the same time, it must avoid collisions with the power lines and masts, for which precise location information is unavailable.

%How far has existing work come? What is the next frontier?
State-of-the-art autonomous systems for power line inspection have two shortcomings:
% (i) Trajectory planning is decoupled from perception and needs accurate information about the location of the power lines and masts; This information is only available in limited cases.
(i) \iros{Control and motion planning are} separated from perception, and they also require precise information regarding the location of power lines and masts. This information is only available in limited cases.
(ii) Collision avoidance and power line tracking are decoupled, which could lead to losing track of the power lines after successfully avoiding an obstacle.
For further details, we refer the reader to a survey on autonomous, vision-based robotic inspection of power lines~\cite{jenssen2018automatic}.

%What does our paper contribute?
We propose a vision-based, tightly-coupled perception and action solution for
autonomous power line inspection that does not require prior information about the power line infrastructure, \iros{such as the  location of the power lines and masts}.
Our method plans and tracks a trajectory that maximizes the visibility of the power line in the onboard-camera view and, at the same time, can safely avoid obstacles such as the power masts.
%What is the key idea? Insights
We achieve this by developing a perception-aware Model Predictive Controller (MPC)~\cite{Falanga18iros} that includes two perception objectives: one for line tracking and one for collision avoidance. 
Adding multiple, and possibly conflicting, perception objectives in the MPC is a challenging task. 
In particular, the optimization could become infeasible and computationally intractable on resource-constrained platforms such as quadrotors.
We overcome this problem by letting the MPC optimize over the weights of the two perception objectives \emph{online}.

\iros{To detect the power lines, we propose a novel perception module that extends the deep-learning--based object detector in~\cite{redmon2016you} to the case of power line detection.}
% The perception module responsible for line detection and tracking is based on a customized deep-learning--based object detector~\cite{redmon2016you}.
The perception module is trained only on synthetic data and transfers zero-shot to real-world images of power lines without any fine-tuning. 
In this way, we overcome the problem of the limited amount of annotated data for supervised learning.

%What do the experiments say?
We demonstrate our system both on simulated data and on a physical quadrotor platform operating in a mock-up power-line infrastructure. We show that our approach is capable of accurately tracking the power lines and avoiding the power masts starting from a rough reference trajectory. 
We believe that our method will contribute to accelerating the deployment of autonomous drones for power line inspection. 
% In addition, our system could inspire the development of novel perception-aware algorithms for autonomous quadrotor navigation.
% \davide{Our contributions are:}
Our main contributions are:
\begin{itemize}
    \item A novel system that tightly couples perception and action for autonomous, vision-based power line inspection.
    \item A model predictive controller that optimizes online the weight of the line tracking and collision avoidance objectives.
    \item \iros{A learning-based power line detector that is trained only on synthetic data and transfers zero-shot to real-world images of power lines.}
    \item Thorough validation of the full system and all its building blocks both in simulation and in the real world on a mock-up power line infrastructure.
    % \item We plan to release the code and dataset upon acceptance of the paper.
\end{itemize}
\section{Related Work} 
An overview of prior works in aerial power line inspection using drones is in~\cite{jenssen2018automatic}.
Planning and control strategies are presented in~\cite{takaya2019development,hui2017novel}.
A PID controller to control the position and orientation of a quadrotor in relation to the power lines is proposed in~\cite{takaya2019development}. 
The solution proposed in~\cite{hui2017novel} uses perspective relation and estimation of the position of the next tower to guide the drone.
Both works loosely couple perception, planning, and control and consequently either need to have access to an accurate reference trajectory or could result in poor line tracking after the collision avoidance maneuver. 
A number of works~\cite{nasseri2018power,tian2015power, gerke2014visual, azevedo2019lidar, dietsche2021powerline} focus on the perception task of detecting and tracking the power lines.
Model-based approaches, such as variants of Hough transform and filters, using cameras are proposed in~\cite{nasseri2018power,tian2015power, gerke2014visual}.
In spite of their high weight and computational load, Lidars could also be employed in some specific situations as proposed in~\cite{azevedo2019lidar}.
Event camera~\cite{gallego2020event} is a novel sensor that provides lower latency and higher dynamic range measurements than standard cameras.
Combining events and standard frames can make robotic perception more robust against motion blur and low light conditions~\cite{sun2021autonomous}.
In~\cite{dietsche2021powerline}, a solution is proposed to detect and track the power lines using event data.

\iros{Extensive literature exists on obstacle avoidance methods for quadrotor flights.
However, state-of-the-art methods~\cite{tordesillas2022panther, wang2021autonomous} cannot directly be applied to the power line inspection problem because they neglect the objective of tracking visual points of interest, such as the power lines.}

Model predictive control~\cite{nguyen2021model} is a powerful solution to couple planning and control for quadrotor autonomous flights.
The benefits of MPC compared to other control strategies are analyzed in~\cite{sun2022comparative}.
MPC has been used for perching on power lines~\cite{paneque2022perception} and agile flights~\cite{romero2022model}.
The first work introducing perception awareness in MPC is~\cite{Falanga18iros} where the authors propose to include a perception objective in the MPC to keep a point of interest in the camera field of view.
Our MPC controller is inspired by~\cite{Falanga18iros}. 
However, we deal with two different perception objectives, line tracking, and collision avoidance, which conflict with each other when the drone approaches the power masts.
\iros{To enable collision avoidance capabilities, in~\cite{schwarm1999chance}, the authors utilized a chance-constrained MPC formulation. This probabilistic collision constraint allows to account for the perceptual uncertainty and consequently enhances the obstacle avoidance robustness.
In~\cite{lin2020robust}, it is shown a real-world application of the chance-constrained MPC for dynamic obstacle avoidance.}
\iros{In~\cite{penin2018vision}, an MPC-based reactive planner for visual target tracking and obstacle avoidance is presented.
Different from our method, this MPC does not directly generate control commands but planned trajectories, which are tracked by another low-level controller.}

\section{Methodology} \label{sec:Methodology}

\subsection{Notation}

\begin{figure}[t!]
    \centering
    \tdplotsetmaincoords{65}{30}
\begin{tikzpicture}[tdplot_main_coords, >=latex]
\tikzset{RPY/.style={x={(\nxx,\nxy)},y={(\nyx,\nyy)},z={(\nzx,\nzy)}}}
\def\propz{0.5}
\def\propx{2.4}
\def\propy{2.2}
\def\r{1}
\def\l{1.2}
\def\f{0.8}
\def\c{0.8}
\def\x{0}
\rotateRPY{-10}{20}{0}
\begin{scope}[RPY]
% frame
\draw[] (-\propx-\x,0,0) -- (\propx-\x,0,0);
\draw[] (-\x,-\propy,0) -- (-\x,\propy,0);
% rotors
\draw[draw=none, fill=black, opacity=0.15] (0-\x,\propy,\propz) circle (\r);
\draw[draw=none, fill=black, opacity=0.15] (\propx-\x,0,\propz) circle (\r) ;
\draw[draw=none, fill=black, opacity=0.15] (0-\x,-\propy,\propz) circle (\r) ;
\draw[draw=none, fill=black, opacity=0.15] (-\propx-\x,0,\propz) circle (\r) ;
\draw (-\x,\propy,0) --++ (0,0,\propz) node[right] {4};
\draw (\propx -\x,0,0) --++ (0,0,\propz) node[right] {2};
\draw (-\x,-\propy,0) --++ (0,0,\propz) node[right] {1};
\draw (-\propx -\x,0,0) --++ (0,0,\propz) node[left] {3};
% forces
\draw[thick,->] (-\x,\propy,\propz) -- ++(0,0,\f);
\draw[thick,->] (\propx-\x,0,\propz) -- ++(0,0,\f);
\draw[thick,->] (-\x,-\propy,\propz) -- ++(0,0,\f);
\draw[thick,->] (-\propx-\x,0,\propz) -- ++(0,0,\f);    

% propeller rotation
\def\psi{60}
\def\xa{30}
\def\xb{0}
\def\xc{-90}
\def\xd{210}
\path (\r-\x,\propy,\propz) arc (0:\xa:\r) coordinate (a);
\draw [->] (a) arc (\xa:\xa-\psi:\r) [draw=black] ;
\path (\propx-\x,\r,\propz) arc (90:\xb:\r) coordinate (b);
\draw [->] (b) arc (\xb:\xb+\psi:\r) [draw=black] ;
\path (\r-\x,-\propy,\propz) arc (0:\xc:\r) coordinate (c);
\draw [->] (c) arc (\xc:\xc-\psi:\r) [draw=black] ;
\path (-\propx-\x,\r,\propz)  arc (90:\xd:\r) coordinate (d);
\draw [->] (d) arc (\xd:\xd+\psi:\r) [draw=black] ;
% quad frame
\rotateRPY{0}{0}{45}
\begin{scope}[RPY]
\draw[->,color=red,text=black] (-\x,0,0) -- ++ (\l,0,0) node[above] {$\bm x_\bfr$};
\draw[->,color=green,text=black] (-\x,0,0) -- ++ (0,\l,0) node[above] {$\bm y_\bfr$};  
\draw[->,color=blue,text=black] (-\x,0,0) -- ++ (0,0,\l) node[left] {$\bm z_\bfr$};     
\rotateRPY{0}{25}{0}

% camera frame
\begin{scope}[RPY, xshift=3.5cm, yshift=-0.75cm]
\draw[->,color=red,text=black] (-\x,0,0) -- ++ (\l,0,0) node[left=1] {$\bm x_C$};
\draw[->,color=green,text=black](-\x,0,0) -- ++ (0,\l,0) node[above] {$\bm y_C$};  
\draw[->,color=blue,text=black] (-\x,0,0) -- ++ (0,0,\l) node[left=1] {$\bm z_C$}; 
\end{scope}

%draw the angle arc
\draw[densely dotted](4.6,0,0) arc (0:-45:0.8) node[right=3] {$30^{\circ}$};

% connect center of body and camera
\draw[dotted]{(0, 0, 0) --++ (5.3, 0, 0)};

\end{scope}
\end{scope}

% world frame
\begin{scope}[xshift=-3cm, yshift=-2cm]
\draw[->,color=red,text=black] (-\x,0,0) -- ++ (\l,0,0) node[right] {$\bm x_\wfr$};
\draw[->,color=green,text=black] (-\x,0,0) -- ++ (0,\l,0) node[above] {$\bm y_\wfr$};  
\draw[->,color=blue,text=black] (-\x,0,0) -- ++ (0,0,\l) node[left] {$\bm z_\wfr$};

% draw a power line
\draw[->]{(0, 0, -1) --++ (8.5, 0, 0)} node[right=3]{};
\node at (5.5, 0, -0.25) [canvas is xz plane at y=0] {Power line detection};

% draw two points
\rotateRPY{0}{0}{0}
\draw[fill=black](1, 0, -1) circle (0.15) node[below=2] {$\bm{p}_{WL_1}$};
\draw[fill=black](7.5, 0, -1) circle (0.15) node[below=2] {$\bm{p}_{WL_2}$};
\end{scope}
\draw{(0, 0, -3.32+0.2-0.7) --++ (0.2, 0, 0)};
\draw{(0.2, 0, -3.32+0.2-0.7) --++ (0, 0, -0.2)};
\draw[dotted]{(0, 0, 0) --++ (0, 0, -4)};
\draw[thick,->,color=black,text=black] (-\x,0,0) -- node[below right] {$\bm g_\wfr$} ++ (0,0,-2*\f);

% Yaw
% \draw[thin,->] (-1.0,-1,-0.5) arc (0:45:0.5) node[midway, right] {$\heading$};
% gravity
\end{tikzpicture}
    \caption{Diagram of quadrotor and power line model.}
    \label{fig:reference_frames}
    \vspace{-1.5ex}
\end{figure}
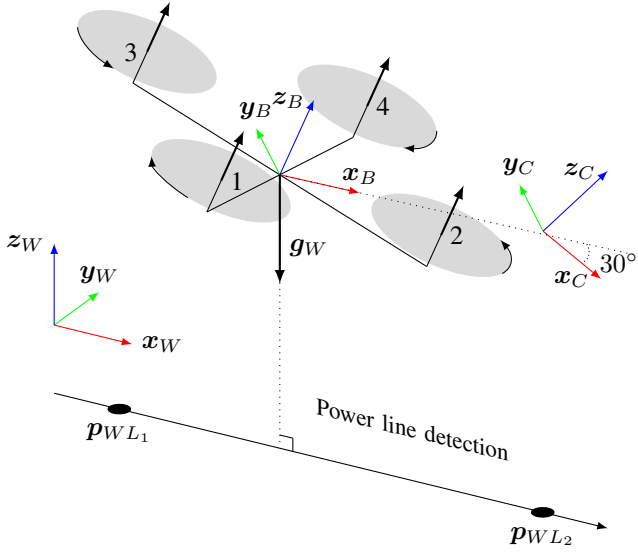

In this manuscript, we define three reference frames.
$W$ is the fixed world frame, whose $z$ axis is aligned with the gravity, $B$ is the quadrotor body frame, and $C$ is the camera frame.
These reference frames are depicted in Fig.~\ref{fig:reference_frames}.
We represent vectors and matrices as bold quantities.
We use capital letters for matrices.
Vectors have a suffix representing the frame in which they are expressed and their endpoint.
For example, the quantity $\bm{p}_{WB}$ represents the position of the body frame $B$ with respect to the world frame $W$.
We use the symbol $\bm{R}_{WB}$ to denote the rotation matrix that rotates a vector from the frame $B$ to the frame $W$.
We use $\bm{q}_{WB}$ to denote the quaternion representation of this rotation.
The time derivative of a vector $\bm{v}$ is represented by $\dot{\bm{v}}$.
In the case of quaternion, the time derivative is defined as $\dot{\bm{q}} = \frac{1}{2} \Lambda (\bm{\omega})$, where $\Lambda(\bm{\omega})$ is the screw-symmetric matrix of the vector $\bm{\omega}$.
The symbol $\odot$ represents the quaternion-vector product.
The symbol $\times$ represents the cross product between two vectors.

\subsection{Quadrotor Dynamics}

Let $\bm{p}_{WB}$, $\bm{q}_{WB}$ and $\bm{v}_{WB}$ be the position, orientation, and linear velocity of the quadrotor expressed in the world frame $W$.
Let $\bm{\omega}_{B}$ be the angular velocity of the body expressed in the body frame $B$.
Additionally, let $c = \Sigma_i c_i$ be the body collective thrust, where $c_i$ is the thrust produced by the i-th motor, $\bm{c} = \begin{bmatrix} 0,  0, c\end{bmatrix}^\intercal$ be the collective thrust vector, $m$ be the mass of the quadrotor, and $\bm{g}_W$ be the gravity vector.
Finally, let $\bm{J}$ be the diagonal moment of inertia matrix and $\bm{\tau}_B$ the body collective torque.
The quadrotor dynamical model is:
\begin{equation}
	\begin{aligned}
		\dot{\bm{x}} = \begin{bmatrix}
			\dot{\bm{p}}_{WB} \\
			\dot{\bm{q}}_{WB} \\
			\dot{\bm{v}}_{WB} \\
			\dot{\bm{\omega}}_{B} \\
		\end{bmatrix} = \begin{bmatrix}
		\bm{v}_{WB}\\
		\frac{1}{2} \Lambda({\omega}_B) \cdot \bm{q}_{WB}\\
		\bm{q}_{WB} \odot \bm{c}/m + \bm{g}_W\\
		\bm{J}^{-1}(\bm{\tau}_B - \bm{\omega}_B\times \bm{J} \cdot \bm{\omega}_B)
	\end{bmatrix}
	\end{aligned}\label{eq:dynamics}
\end{equation}
The body torque $\bm{\tau}_B$ is represented by:
\begin{equation}
	\bm{\tau}_B = \begin{bmatrix}
		-d_{x_0}&-d_{x_1}&d_{x_2}&d_{x_3}\\
		d_{y_0}&-d_{y_1}&-d_{y_2}&d_{y_3}\\
		-c_{\tau}&c_{\tau}&-c_{\tau}&c_{\tau}
	\end{bmatrix}\begin{bmatrix}
		c_1\\c_2\\c_3\\c_4
	\end{bmatrix}
\end{equation}
where $d_{x_i}$, $d_{y_i}$ for $i$ = [1, 2, 3, 4] are the distances of each rotor $i$ to the body frame and $c_\tau$ is the rotor drag constant.
The state and input vector of the system are $\bm{x} = [\bm{p}_{WB}^\intercal, \bm{q}_{WB}^\intercal, \bm{v}_{WB}^\intercal, \bm{w}_{B}^\intercal]^\intercal$ and  $\bm{u} = [c , \bm{\omega}_B^\intercal]^\intercal$.

\subsection{MPC Formulation}\label{ch2_mpc}
The system dynamics in Eq.~\ref{eq:dynamics} can be written in compact form as $\dot{\bm{x}} = f(\bm{x}, \bm{u})$.
We compute the discrete-time version of it by using a Runge-Kutta method of 4th order with time step $dt$: $\bm{x}_{i+1} = f(\bm{x}_i, \bm{u}_i, dt)$.
The MPC formulation is a non-linear program with quadratic costs:
\begin{equation}
	\begin{aligned}
		\mathcal{L}_{org} =\, &\bar{\bm{x}}_N^\intercal \mathcal{Q}_{x,N} \bar{\bm{x}}_N ~+ \sum_{i=0}^{N-1} \left(\bar{\bm{x}}_i^\intercal \mathcal{Q}_{x} \bar{\bm{x}}_i + \bar{\bm{u}}_i^\intercal \mathcal{R} \bar{\bm{u}}_i\right) \\
		& \argmin_{\bm{u}} \; \mathcal{L}_{org} \\
		& \text{s.t.} 
		\begin{array}[t]{l}
			\bar{\bm{x}}_0 = \bm{x}_{init}\\
			\bm{x}_{i+1} = f(\bm{x}_i, \bm{u}_i)\\
        \bm{u}_{\textrm{min}} \leq  \bm{u}_i\leq\bm{u}_{\textrm{max}}.
		\end{array}
	\end{aligned}\label{eq:classical_mpc}
\end{equation}
This is solved as a sequential quadratic program (SQP) executed in a real-time iteration scheme~\cite{diehl2006fast}.
The values $\bar{\bm{x}} = \bm{x} - \bm{x}_s$ and $\bar{\bm{u}} = \bm{u} - \bm{u}_s$ refer to the difference with respect to the reference of each value.
We implement this optimization problem in ACADO~\cite{houska2011acado} and use the qpOASES solver~\cite{ferreau2014qpoases}.
\begin{figure}
	\centering
	\begin{tikzpicture}[x=0.018cm,y=0.018cm]
\definecolor{my_green}{rgb}{0.1, 0.5, 0.1}
\definecolor{my_blue}{rgb}{0.1, 0.4, 0.9}

% Rectangle
\node (rect) at (0, 0) [draw, minimum width=5.4cm, minimum height=3.6cm] {};
\draw [] (rect) node[above, yshift=1.8cm] {Image Frame};
\foreach \i in {-5 ,...,5} {
        \draw [line width=0.05,gray] (\i * 30, -100) -- (\i * 30,100)  node [below] at (\i * 30,100) {};
        }

\foreach \i in {-3 ,...,3} {
    \draw [line width=0.05,gray, opacity=0.5] (-150, \i * 30) -- (150, \i * 30)  node [below] at (150, \i * 30) {};}
% Green middle line
\draw [color=my_green, line width=1.2] (0, -100) -- (0, 100) ;

% Blue middle line
\draw [color=my_blue, line width=1.2][fill={rgb, 255:red, 65; green, 117; blue, 5 }  ,fill opacity=1 ]   (100 - 5*11, -100 + 5 * 11 - 30) -- (-100 + 5*3, 100 - 5 * 3 - 30) ;

%  Center circles
\draw [color=my_green, fill=my_green](0, 0) circle (5) ;
\draw [color=my_blue, fill=my_blue] (-30, -0) circle (5) ;

% Blue dashed line
\draw [dotted][color=my_blue, line width=1.2] (100 - 5 * 11, -100 + 5 * 11  - 30) -- (70, -100) ;

\draw [dotted][color=my_blue, line width=1.2](-100 + 5*3, 100 - 5*3 - 30) -- (-130, 100) ;

% Connect two center circles
\draw [line width=0.5](0.25, -0.5) -- (-35, 35) ;

% Connect two center points
\draw [line width=0.5](-30, -0) -- (0, 0);

% Draw arrows
\draw [line width=0.5][stealth-](-30, 30) -- (-30 + 20,  30 + 20);
\draw [line width=0.5][stealth-](-45, 15) -- (-45 - 20, 15 - 20);

% Connect two arrows
\draw [line width=0.5](-30, 30) -- (-45, 15);
% Perpendicular symbol
\draw [line width=0.1](-30 + 4, 30 - 4) -- (-30, 30 - 8);
\draw [line width=0.1](-30 - 4, 30 - 4) -- (-30, 30 - 8);

% Draw vertical line
\draw [line width=0.5](-30, 0) -- (-30, -10 - 80);
\draw [line width=0.5][stealth-](0, -10 - 80  +5 ) -- (0 + 25, - 10 - 80 + 5); 
\draw [line width=0.5][stealth-](-30, -10-80  +5 ) -- (-30 - 25, -10-80 + 5); 
\draw [line width=0.5](-30, -10-80 + 5) -- (0, -10-80 + 5);

%Draw angle arc
\draw (0,-45) arc (270:315:15);

% Draw texts
\draw [color=my_green] (65, 90) node[below] {Reference line};
\draw [color=my_blue] (70, -8) node[below] {Detected line};
\draw (-40, 50) node[below] {$r$};
\draw (-50, -60) node[below] {$h$};
\draw (10, -40) node[below] {$\theta$};

\end{tikzpicture}
	\vspace{0.5ex}
	\caption{Illustration of a reference and detected line onto the image frame.}
	\vspace{-1.5ex}
	\label{fig:lineeval}
\end{figure}
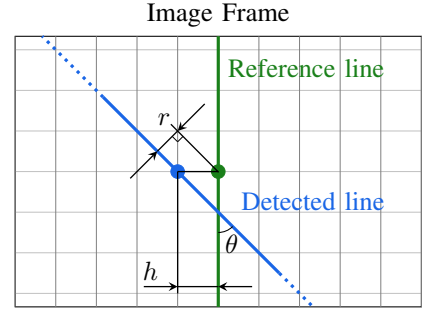

\subsection{Perception Objectives}\label{ch2_perception}
We include two perception objectives: one for line tracking and one
for collision avoidance, in the MPC formulation proposed in Eq.~\ref{eq:classical_mpc}.

\medskip
\textbf{Line Tracking: }
The purpose of this objective is to keep the power line in the center of the image (c.f. Fig.~\ref{fig:lineeval}), to maximize data quality for visual inspection, \iros{and to keep a safe distance from the power lines}.
We derive the line tracking objective in this section by extending the perception objective proposed for a single point in~\cite{Falanga18iros}.
We denote the position of the line endpoints in the world frame $W$ as $\bm{p}_{W{L_j}} \; j\in\left\{1, 2\right\}$.
These points are transformed to the camera frame $C$ by:
\begin{equation}
	\bm{p}_{C{L_j}} = (\bm{q}_{WB} \cdot \bm{q}_{BC})^{-1} \odot (\bm{p}_{W{L_j}} - (\bm{q}_{WB} \odot \bm{p}_{BC} + \bm{p}_{WB}).
	\label{eq:line_endpoints_camera_frame}	
\end{equation}
The points: $\bm{p}_{C{L_1}}$ and $\bm{p}_{C{L_2}}$ are projected into the image plane coordinates: $[u_1, v_1], [u_2, v_2]$ according to the classical pinhole camera model~\cite{Szeliski10book}.
The cartesian coordinates are transformed into the polar coordinates as:
\begin{equation}\label{eq:cartesian_to_polar}
		\theta = \arctan\left(-\frac{u_2 - u_1}{v_2 - {v_1}}\right), \;
		r = \left(v_1 - \frac{v_2 - v_1}{u_2 - u_1} u_1 \right) \sin\theta.
\end{equation}
We introduce a new variable $\bar{\bm{z}}$ in our MPC formulation:
\begin{equation}
	\bm{z} = \left(\begin{array}{c}
		\theta \\
		r \\
		d
	\end{array} \right), \qquad
	\bm{z}_s = \left(\begin{array}{c}
		0 \\
		0 \\
		d_s
	\end{array} \right), \qquad
	\overline{\bm{z}} = \bm{z} - \bm{z}_s.
\end{equation}

The variable $d$ represents the distance of the line to the body frame (c.f. Fig.~\ref{fig:reference_frames}) and $d_s$ represents the \iros{target} value of $d$. \iros{The value of $d_s$ is set by the user according to the desired distance of the flight from the power lines.} 
\medskip
\textbf{Obstacle Avoidance:}
Inspired by~\cite{zhu2019chance,lin2020robust}, we include collision avoidance capabilities in our MPC by means of a collision cost and a collision constraint.
The collision cost $l_{o}$ is formulated with the logistic function:
\begin{equation}\label{eq:collision_cost}
	l_{o} = \mathcal{Q}_o/(1 + \text{exp}\!\left(\lambda_o\left(d_{o} - r_o\right)\right)),
\end{equation}
where $d_{o}$ represents the norm of the distance of the body frame to the detected obstacle.
The values ${Q}_o, \lambda_o, r_o$ are constant quantities that represent weight, smoothness, and distance threshold, respectively.\\
The collision constraint is formulated as a probabilistic chance constraint to account for the uncertainty in the drone state and in the obstacle detection.
The objective of this constraint is to ensure that the probability of the collision with an obstacle is less than a predefined threshold: $\text{Pr}\{C_{o}\} < \delta$.
We model obstacles as ellipsoids. 
% This is a convex representation that is suitable for modelling power masts.
Let $a_{o}, b_{o}, c_{o}$ be the semi-principal axes of the ellipsoid modeling an obstacle, and $r$ the radius of a safety area around the quadrotor body frame.
The quadrotor is considered to be in collision with the obstacle when:
\begin{equation}
C_{o} : \left(\mathbf{p}_{WB} - \mathbf{p}_{WO}\right)^\intercal\bm{\Omega}_{o}\left(\mathbf{p}_{WB} - \mathbf{p}_{WO}\right) \leq 1,
\end{equation}
where $\bm{\Omega}_{o}$ is the uncertainty matrix defined as $\bm{\Omega}_{o} = \mathbf{R}_{WO}^\intercal\ \cdot \text{diag}\left(\frac{1}{\left(a_{o} + r\right)^2}, \, \frac{1}{\left(b_{o} + r\right)^2}, \, \frac{1}{\left(c_{o} + r\right)^2}\right) \cdot \mathbf{R}_{WO}$.
The quantity \iros{$\mathbf{p}_{WO}$ and} $\mathbf{R}_{WO}$ represent the \iros{position and} orientation of the obstacle with respect to the world frame $W$.
Assuming that the quadrotor and obstacle positions are random variables distributed according to Gaussian distributions: $\mathbf{p}_{WB}\sim\mathcal{N}(\hat{\mathbf{p}}_{WB}, \bm{\Sigma})$, and $\mathbf{p}_{WO}\sim\mathcal{N}(\hat{\mathbf{p}}_{WO}, \bm{\Sigma}_{o})$, respectively, we derive the deterministic form of the chance constraint as:
\begin{align}\label{eq:cc}
    \bm{n}_o^\intercal \bm{\Omega}_{o}^{\frac{1}{2}}(\hat{\bm{p}}_{WB} - \hat{\bm{p}}_{WO}) &- 1 \geq \text{erf}^{-1}(1-2\delta) \cdot \nonumber \\&
    \sqrt{2\bm{n}_o^\intercal \bm{\Omega}_{o}^{\frac{1}{2}} (\bm{\Sigma} + \bm{\Sigma}_o)\bm{\Omega}_{o}^{\frac{1}{2}} \bm{n}_o}
\end{align}
where $\bm{n}$ is the normalized distance from the body frame to the obstacle and erf($x$) is the standard error function for Gaussian distributions~\cite{Barfoot15book}.
We rearrange Eq.~\ref{eq:cc} and write it using the shorthand $cc(\hat{\bm{p}}_{WB}, \, \Sigma_{B}, \, \hat{\bm{p}}_{WO}, \, \Sigma_{O}) \leq  0$ hereafter.

\subsection{Perception-aware MPC for power line inspection}\label{sec:perception_aware_mpc}
The two perception objectives of line tracking and obstacle avoidance conflict when the quadrotor approaches the power masts.
For this reason, finding constant weights to attribute to these objectives in the MPC is difficult.
Our solution is to adapt online these weights.
To this end, we introduce a new state variable, $\alpha$. 
This variable varies in the interval $[0, \alpha_{max}]$.
For values close to 0, high priority is given to line tracking.
On the contrary, for values close to $\alpha_{max}$, high priority is given to collision avoidance.
The perception-aware MPC proposed in this work is:
\begin{equation}
	% \label{eq:CCLAMPC}
		\begin{aligned}
%   \mathcal{L}_{org} =\, &\bar{\bm{x}}_N^\intercal \mathcal{Q}_{x,N} \bar{\bm{x}}_N ~+ \sum_{i=0}^{N-1} \left(\bar{\bm{x}}_i^\intercal \mathcal{Q}_{x,i} \bar{\bm{x}}_i + \bar{\bm{x}}_i^\intercal \mathcal{R}_i \bar{\bm{x}}_i\right) \\
\argmin_{\bm{u}, \alpha}& \; \mathcal{L}_{org}+\mathcal{L}_{per} \\
			\mathcal{L}_{per} =&\,  \sum_{i=0}^{N-1} \left(
			\bar{\alpha}_i^2 \bar{\bm{z} }_{i}^\intercal \mathcal{Q}_{p} \bar{\bm{z} }_{i} +
			l_{o} +\mathcal{Q}_{\alpha} \alpha_i^2\right) \\
			 &\text{s.t.}  
			\begin{array}[t]{l}
				\bar{\bm{x}}_0 = \bm{x}_{init}\\
				\bm{x}_{i+1} = f(\bm{x}_i, \bm{u}_i)\\
				\bm{u}_{\textrm{min}} \leq  \bm{u}_i\leq\bm{u}_{\textrm{max}} \\
				\hspace{-0.65cm}cc\!\left(\hat{\bm{p}}_{WB, i}, \, \Sigma_{B,i}, \, \hat{\bm{p}}_{WO, i}, \, \Sigma_{O,i} \right) + c \bar{\alpha}_i \leq 0 \\
				0 \leq \alpha_i \leq \alpha_{max},
			\end{array}
	\end{aligned}\label{eq:perception_aware_mpc}
\end{equation}
where $\bar{\alpha} = 1 - \alpha/\alpha_{max}$ and $c$ is a constant value \iros{that is used to weight the priority of the chance constraint.}

\subsection{Line Detection and Tracking}
In this work, we propose a \iros{novel} deep-learning-based power line detector based on the object detector~\cite{redmon2016you, glenn_jocher_2022_7002879}.
% The object detector~\cite{glenn_jocher_2022_7002879} outputs bounding boxes around the detected objects.
\begin{figure}[t!]
    \centering
    \includegraphics[width=\columnwidth]{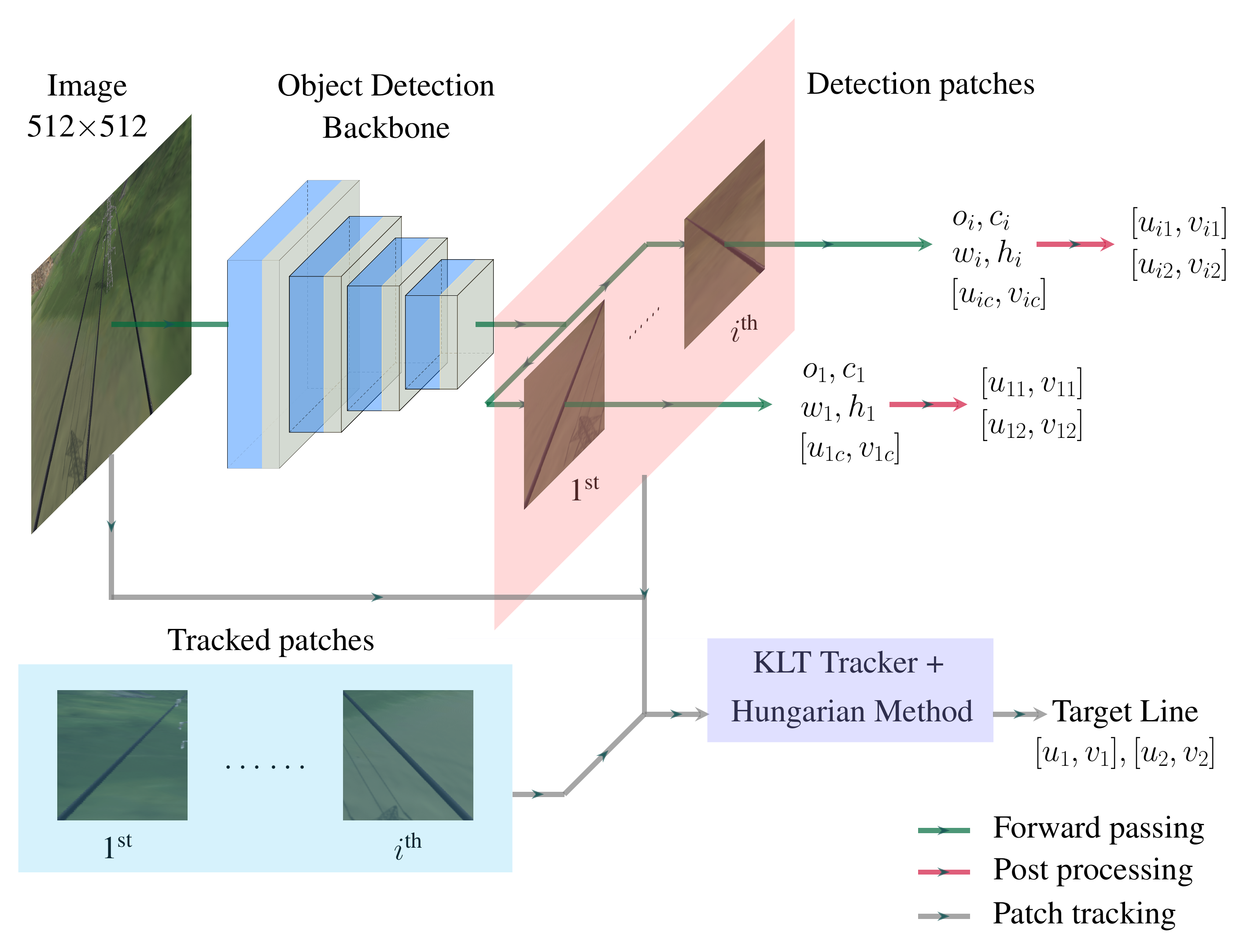}
    \caption{ \iros{Overview of our learning-based power line detector and tracker. The detector takes a single RGB image as input and outputs end points of the detected power lines in pixel coordinates. The center patch of each detection is matched with the prediction of the previous patch using the Hungarian method~\cite{kuhn1955hungarian}. We use a KLT tracker~\cite{lucas1981iterative} to perform tracking. The final output is the tracked lines endpoints which are given to the MPC.}}
    \label{fig:perception}
    \vspace{-1.5ex}
\end{figure}
\iros{
As shown in Fig.~\ref{fig:perception}, our detector takes monocular RGB images as input and outputs: 
(i) width $w_i$ and height $w_i$ of the bounding boxes that fully contain the detected power line, where $i$ indicates the number of detection;
(ii) the inclination of the line $o_i$ (either +1 or -1).
The endpoints of positive inclined lines, i.e., $o_i=+1$, correspond to the top-left and bottom-right corners of the bounding box.
The endpoints of negative inclined lines correspond to the top-right and bottom-left corners of the bounding box;
(iii) the center of the line $[u_{ic}, v_{ic}]$;
(iv) the confidence score $c$ of the prediction. 
If the confidence score is not larger than a predefined threshold (we use 0.8 in all our experiments), the detection is labeled as invalid and is not used.
To perform tracking, we use a tracking-by-detection approach to track the detected power lines.}
In the first step, we track the center patch of the detected bounding box, of dimension 25$\times$25 pix, using a 4 layers Lukas-Kanade tracker (KLT)]~\cite{lucas1981iterative}.
In the second step, we compute the similarity score between bounding boxes in two consecutive frames as the sum of the KLT error and the area difference between the two bounding boxes normalized by the image size.
In the final step, we find the best matches using the 
Hungarian method~\cite{kuhn1955hungarian}. The final outputs are the endpoints of the tracked line in the pixel coordinate $[u_1, v_1], [u_2, v_2]$. Then we transform the endpoints, incorporating depth detection and state estimation, into the world frame to derive $\bm{p}_{W{L_1}}, \bm{p}_{W{L_2}}$. 
% These values are then integrated into our MPC. in~\ref{eq:line_endpoints_camera_frame}.

\begin{figure}[t!]
    \centering
    \includegraphics[ width=0.95\columnwidth]{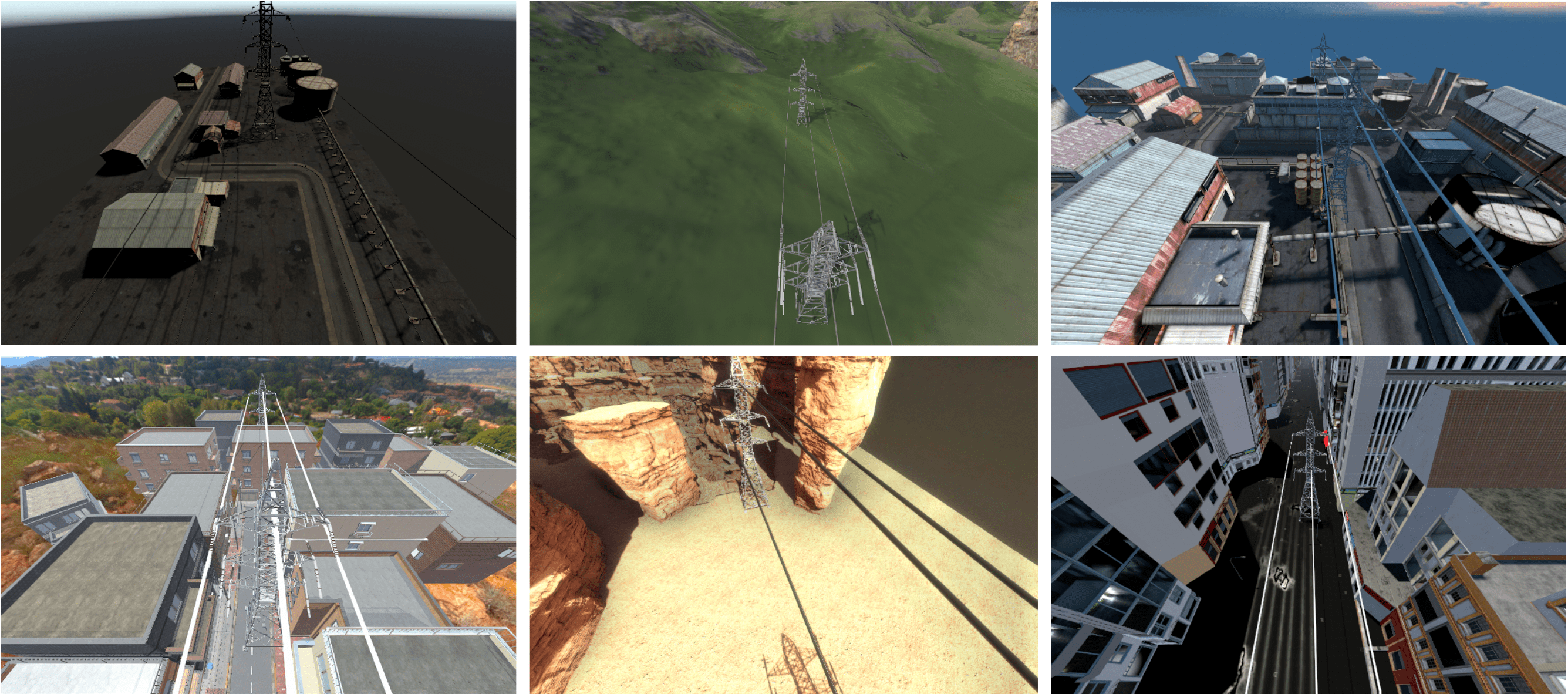}
    \vspace{0.5ex}
    \caption{Sample images from the proposed power line dataset.}
    \label{fig:scenes}
    \vspace{-1.5ex}
\end{figure}

Among the datasets which contain power lines~\cite{zhang2019detecting, abdelfattah2020ttpla}, there is a limited amount of labeled data and scene variability.
Consequently, they are not suitable for training a robust power line detector.
For this reason, we created a new dataset for power line detection based on the photo-realistic simulator Flightmare~\cite{song2021flightmare}. 
We collected a dataset of $\sim$30k images of labeled power lines of different colors and thicknesses in different 6 environments (c.f. Fig.~\ref{fig:scenes}).
The dataset is split into training (80\%), validation (10\%), and test  (10\%) sequences.
We train our power line detector on the training sequence and use the validation sequence for hyperparameters tuning.

\iros{Our current approach requires that the user specifies which power line to track, e.g., the ID $i$ from the detection.
The ID cannot be changed during the flight.
We leave as part of future work the extension of the MPC formulation to track different lines.
A current limitation of the line tracked is the missing information on the temporal history of the lines. 
If a power line is not tracked between two frames, the detector is unable to continue tracking the previous line. 
To address this issue, a possible solution would be to implement a hibernation mechanism as in~\cite{dietsche2021powerline}.}

\subsection{Obstacle Detection}
We use ellipsoids with constant but unknown sizes to represent obstacles.
The state of an obstacle is defined as $\left(\bm{p}_{WO}, a_o, b_o, c_o\right)$, which contains the position of the center of the ellipsoid in the world frame and the length of its semi-principal axes.
Our obstacle detection module is based on U- and V- disparity maps as proposed in~\cite{oleynikova2015reactive}.
The U-map is a histogram of disparity values accumulated over the columns of the image.
The V-map is a histogram of disparity values accumulated over the rows of the image.
These disparity maps are used to detect obstacles from depth images. 

\section{Experiments}\label{sec:Experiments}
In this section, we validate our system by answering the following questions:

\begin{itemize}
    \item Why use our learning-based line detector instead of classical methods?
    \item The initial reference trajectory can be in collision with the power masts. In this case, can our system safely avoid the power masts?
    \item Does tightly-coupled perception \iros{and action} improve data quality for visual inspection?
\end{itemize}
In addition, we demonstrate our system in the real world on a mock-up power line infrastructure.
The proposed perception-aware MPC runs inside the quadrotor control stack~\cite{foehn2022agilicious}.
\medskip

\textbf{Benefits of our learning-based line detector:}
\begin{figure}[t!]
    \centering
    \includegraphics[width=0.98\columnwidth]{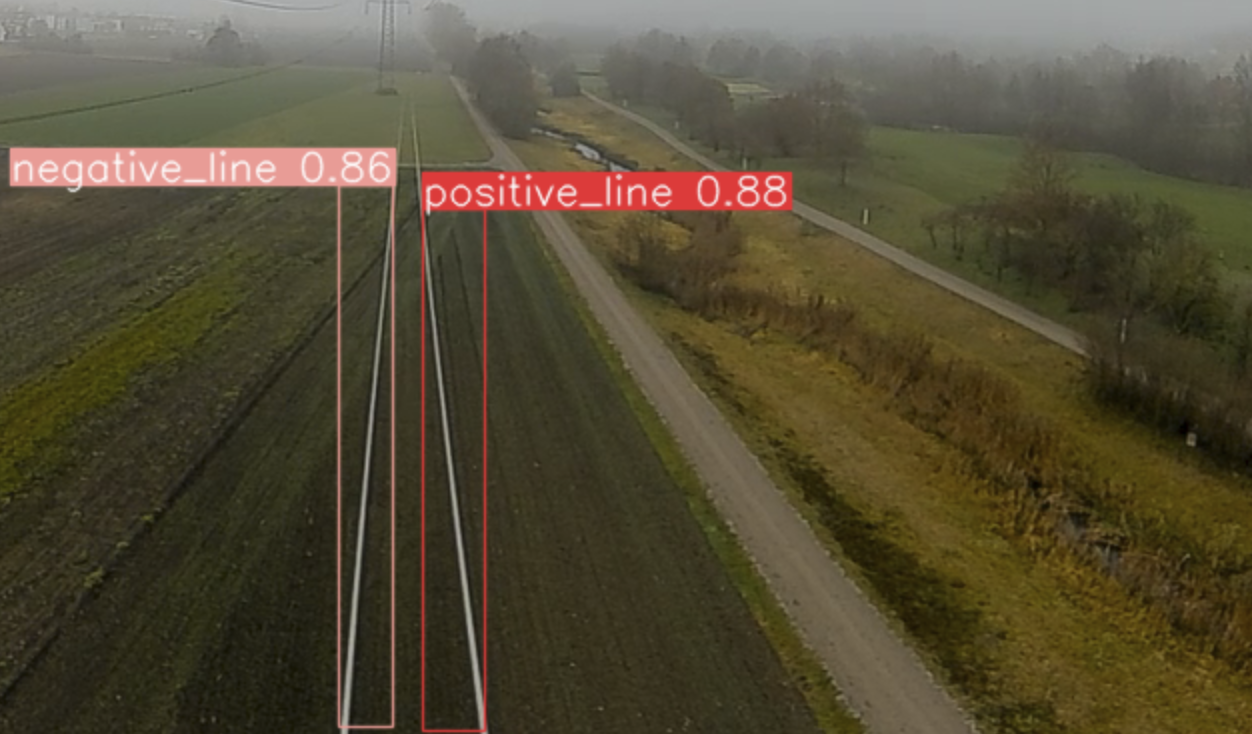}
    \vspace{0.5ex}
    \caption{A sample result of our line detector validated using a real-world image. Our detector shows correct prediction with high confidence in a challenging real-world scene, despite only being trained with synthesized images.}
    \label{fig:yolo}
    \vspace{-2.5ex}
\end{figure}
In these experiments, we compare the proposed learning-based line detector against a traditional line detector approach based on the Hough transform algorithm~\cite{illingworth1988survey}.
We design a traditional line detection baseline that uses the Canny edge detector algorithm~\cite{bao2005canny} to detect edge features in the image and the Hough transform algorithm to estimate the line parameters.
We also design a probabilistic version of this algorithm, Probabilistic Hough Transform (P-Hough), that runs on a sampled subset of the detected edges.
The dimension of this subset depends on predefined thresholds that vary according to the number of detected edges.
The parameters of the traditional approach were tuned on the training sequences of our simulated power line dataset.
The results on the test sequences are listed in Table~\ref{tab:quant_eval_in}.
We use the metrics proposed in~\cite{zhao2021deep}, which are the Precision, Recall, and the F1 score of the line matching results based on Chamfer distance and EA score~\cite{zhao2021deep}.
Our line detector greatly outperforms the traditional approach.
Generally, it is difficult to find a set of parameters for the traditional approach that generalizes to the different environments of our dataset.
On the contrary, our learning-based detector is able to generalize to all the environments in the dataset.
Furthermore, we demonstrate that our proposed line detector is able to generalize to real-world data without fine-tuning.
Fig.~\ref{fig:yolo} shows the output of the line detector on a real-world power line image recorded onboard a quadrotor. 
We refer the reader to the accompanying video where we include the results of our line detector on the full sequence of images recorded onboard a quadrotor flying above the real-world power lines.
\begin{table}[t!]
	\centering
	\begin{tabular}{lcccccccl}
		\toprule
		\textbf{Method}&\multicolumn{3}{c}{\textbf{Chamfer Distance}} & \multicolumn{3}{c}{\textbf{EA Score}}\\
		\midrule
		&P&R&F&P&R&F\\
		Hough&0.70&0.32&0.44&0.46&0.19&0.28\\
		P-Hough&0.26&0.31&0.28&0.10&0.16&0.10\\
		\midrule
		\textbf{Ours}&\textbf{0.92}&\textbf{0.77}&\textbf{0.84}&\textbf{0.93}&\textbf{0.78}&\textbf{0.85}\\
        \midrule
		Improvement (\%)&31&141&91&102&311&204\\
		\bottomrule
	\end{tabular}
    \vspace{1ex}
	\caption{Quantitative evaluation of the performance of the proposed learning-based line detector and of the classical approaches.}

	\label{tab:quant_eval_in}
      \vspace{-2.5ex}

\end{table}
We also evaluate the line detection algorithms in terms of the time they need to process a single frame, which we name running time.
Such a running time was computed on a laptop equipped with an Intel Xeon E3-1505M v5 (2.80GHz) CPU and Nvidia Quadro M2000M GPU, and on Nvidia Jetson TX2, which is the computer available onboard our quadrotor platform.
The results are shown in Table~\ref{tab:trad_methods}. 
\iros{
Furthermore, to justify the need for the proposed line detector, we compare our approach against the state-of-the-art models~\cite{pan2018spatial, tabelini2021keep, liu2021condlanenet, li2019line, chen2019pointlanenet} for pixel-wise line detection regarding the computation time.
\begin{table}[b]
	\centering
	\begin{tabular}{lcccl}
		\toprule
		\textbf{Method}  & \textbf{Run Time [ms]} & \textbf{GFlops}  \\
		\midrule
		SCNN~\cite{pan2018spatial} & 133.33 & 328.4 \\
		LaneATT~\cite{tabelini2021keep} & 38.46 & 70.5\\
		CondLaneNet~\cite{liu2021condlanenet} & 17.18 & 44.8\\
		Line-CNN~\cite{li2019line} & 27.93 & - \\
		PointLaneNet~\cite{chen2019pointlanenet} & 14.08&-\\
		\midrule
		\textbf{Ours} & \textbf{1.80} & \textbf{0.0042}\\ 
		\bottomrule
	\end{tabular}
 \vspace{1ex}
	\caption{\iros{Comparison of the state-of-the-art line detection methods with respect to their inference speeds and operation numbers. Our approach outperforms the state-of-the-art methods in terms of computational efficiency, All experiments use the same input image size of $512\times512$ on Nvidia GeForce RTX 2080Ti GPU.}}
	\label{tab:learn_methods}
  \vspace{-3.5ex}

\end{table}
As shown in Table \ref{tab:learn_methods},
the high latency of these models makes them unsuitable for real-time deployment onboard resource-constrained platforms such as quadrotors. 
The high latency is mainly because these methods predict pixel-wise segmentation, which is not needed in our use case.}\\

\begin{figure}[t!]
    \centering
    \vspace{-1ex}
    \begin{tikzpicture}
            \node[inner sep=0pt] (pic) at (0,0){\includegraphics[width=1.0\columnwidth]{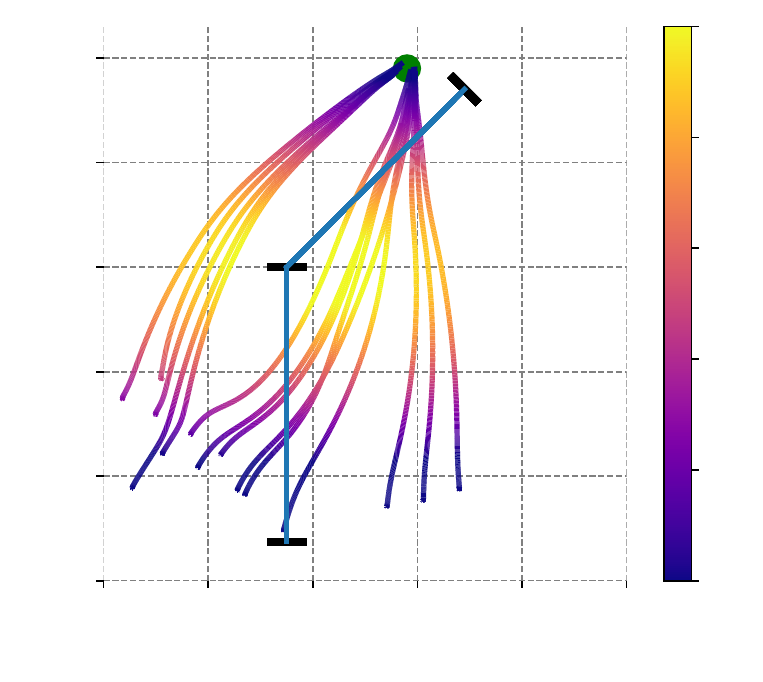}};
            % x coodinate
            \node[inner sep=0pt] at (-3.15, -3.1){0};
            \node[inner sep=0pt] at (-1.96, -3.1){2};
            \node[inner sep=0pt] at (-1.96+1.19, -3.1){4};
            \node[inner sep=0pt] at (-1.96+1.19*2, -3.1){6};
            \node[inner sep=0pt] at (-1.96+1.19*3, -3.1){8};
            \node[inner sep=0pt] at (-1.96+1.19*4, -3.1){10};

            \node[inner sep=0pt] at (-1.96+1.19*1.5, -3.5){$x$ [m]};

            % y coordinate
            \node[inner sep=0pt] at (-3.5, -2.75){0};
            \node[inner sep=0pt] at (-3.5, -2.75 + 1.19){2};
            \node[inner sep=0pt] at (-3.5, -2.75 + 1.19*2){4};
            \node[inner sep=0pt] at (-3.5, -2.75 + 1.19*3){6};
            \node[inner sep=0pt] at (-3.5, -2.75 + 1.19*4){8};
            \node[inner sep=0pt] at (-3.5, -2.75 + 1.19*5){10};
            \node[inner sep=0pt, rotate=90] at (-3.9, -2.75 + 1.19*2.5){$y$ [m]};

            % colorbar
            \node[inner sep=0pt] at (4, -2.7) {0.0};
            \node[inner sep=0pt] at (4, -2.7+1.25) {0.2};
            \node[inner sep=0pt] at (4, -2.7+1.25*2) {0.4};
            \node[inner sep=0pt] at (4, -2.7+1.25*3) {0.6};
            \node[inner sep=0pt] at (4, -2.7+1.25*4) {0.8};
            \node[inner sep=0pt] at (4, -2.7+1.25*5) {1.0};

            %ABC
            \node[inner sep=0pt] at (-0.5, -2.3) {\large A};
            \node[inner sep=0pt] at (-1.45, 0.85) {\large B};
            \node[inner sep=0pt] at (1.4, 2.9) {\large C};

            %Goal
            \node[inner sep=0pt] at (-0.2, 3.5) {Goal};

    \end{tikzpicture}
    \caption{Visualization of the flown trajectories color-coded according to the values of $\alpha$, c.f. Sec.~\ref{sec:perception_aware_mpc}.
    Our MPC is able to find collision-free trajectories starting from non-collision-free reference trajectories.}
    \label{fig:sim}
\end{figure}
\noindent
\textbf{Robustness to unknown location of the power masts: }
\begin{table}[t!]
	\centering
	\begin{tabular}{lcccl}
		\toprule
		\textbf{Method}  & \textbf{Run Time Laptop [ms]} & \textbf{Run Time TX2 [ms]} \\
		\midrule
		Hough &7.19&18.52\\
        \textbf{P-Hough}  &\textbf{4.61}&\textbf{16.13}  \\
		\midrule
		Ours& 4.69& 20.83\\ 
		\bottomrule
	\end{tabular}
    \vspace{1ex}
	\caption{Running time, i.e. time required to process a single image, of the proposed learning-based line detection algorithm and of the classical approaches.}
    \vspace{-3.5ex}
	\label{tab:trad_methods}
\end{table}
Generally, the location of the power masts is either unavailable or GPS coordinates with errors up to several meters ($>$ 10) are available.
For this reason, it is not possible to plan a collision-free reference trajectory.

In this experimental setting, we evaluate the performance of our system in the case where the initial reference trajectory is in collision with the power masts.
We run our tests within the Flightmare simulation environment and assume that the MPC has access to the ground truth obstacle position and only vary the reference trajectory.
We designed an environment with 3 power masts (labelled A, B, C) as shown in Fig.~\ref{fig:sim}.
We randomly sampled 100 starting points in a rectangular region of size 8$\times$3 m in the vicinity of the power mast A (the center of A is located in the middle of the bottom edge of this rectangular region).
The endpoint is fixed 1 m away from the left side of the power mast C.
The reference trajectory given to the MPC is a straight line connecting the starting and end points with no yaw change.
Some of these reference trajectories are in collision with the power mast B.
Our system achieves a 100\% success rate, i.e. no collision with the power masts.
We show in Fig.~\ref{fig:sim}, 13 flown trajectories.\\

\noindent
\textbf{Benefits of tightly-coupling perception and action:}
In this experiment, we evaluate the benefits of tightly-coupling perception, planning, and control in terms of visibility of the power lines compared to the classical MPC formulation (c.f. Eq.~\ref{eq:classical_mpc}).
We use the proposed learning-based line detector and assume that the MPC has access to the ground-truth position of the power masts.
We use the similarity score $\mathcal{S}$ as the evaluation metric, which is defined as:
\begin{equation}
	\mathcal{S}_\theta = 1- \frac{2\theta}{\pi},\;
	\mathcal{S}_d = 1 - \frac{h}{\sqrt{w^2 + l^2}},\;
	\mathcal{S} = (\mathcal{S}_\theta \cdot \mathcal{S}_d)^2
 \label{eq:sim_score}
\end{equation}
\noindent
where $\theta$ is the normalized angular distance between the detected and the reference line, and $h$ is the normalized distance between the center points of the lines (c.f. Fig.~\ref{fig:lineeval}).
The quantity $h$ is normalized by the length of the image diagonal, where $w$ and $l$ are the image width and height, respectively.
We run the experiments on the testing sequences of our dataset which are from 4 different environments: Simple, Warehouse, Forest, and Village.
The results are in Table~\ref{tab:eval}. 
Our perception-aware MPC improves the line visibility on average by 40\%.

In addition, we evaluated the run time of our perception-aware MPC formulation (c.f. Eq.~\ref{eq:perception_aware_mpc}), the classical MPC formulation, an MPC including the only line tracking objective (Tracking MPC), and, an MPC including the only collision avoidance objective (Avoidance MPC).
We report the update and solver time when running the algorithms on an Nvidia Jetson TX2 platform, which is the computer onboard our quadrotor.
The update time is the period of time between two consecutive control commands.
The solver time is the period of time taken by the solver to optimize the MPC objective.
The results are shown in Table~\ref{tab:mpc_run_time}.
Our proposed MPC requires more computational time than the classical formulation mainly due to the collision objective.
However, the update is fast enough to deploy the proposed MPC on a real quadrotor.
\begin{figure}[t!]
    \centering
    \includegraphics[width=0.95\columnwidth]{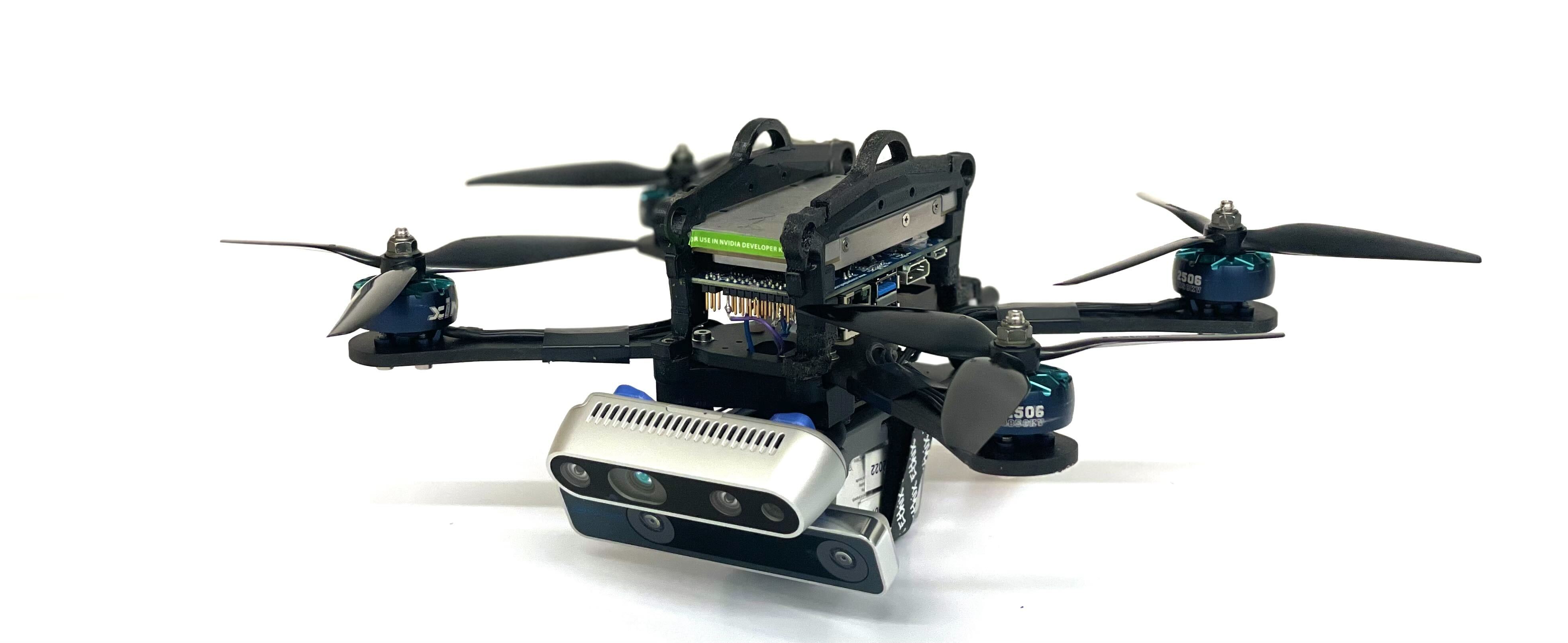}
    \caption{Quadrotor used in real-world experiments.}
    \label{fig:frame}
    \vspace{-2ex}
\end{figure}

\begin{table}[t!]
	\centering
	\begin{tabular}{lccccccl}
		\toprule
		&\textbf{Simple}  & \textbf{Warehouse} & \textbf{Forest} & \textbf{Village}& \textbf{Mean}      
  \\
  \midrule
		\textbf{Our MPC}  & \textbf{0.74} & \textbf{0.63} & \textbf{0.61} & \textbf{0.67} & \textbf{0.67}\\
		Classical MPC  & 0.53 & 0.49 & 0.44 & 0.45 & 0.48\\
		\bottomrule
	\end{tabular}
	\vspace{1ex}
	\caption{Quantitative comparison of the proposed perception-aware MPC (Eq.~\ref{eq:perception_aware_mpc}) against the classical MPC formulation (Eq.~\ref{eq:classical_mpc}). The evaluation metric refers to the visibility of the line (Eq.~\ref{eq:sim_score}).}
	\vspace{-2.5ex}
	\label{tab:eval}
\end{table}

\begin{table}[b!]
	\centering
	\begin{tabular}{lcccl}
	\toprule
	&\textbf{Update Time [ms]}&\textbf{Solver Time [ms]}\\
	\midrule
	Classical MPC&0.88$\pm$0.27 & 0.65 $\pm$0.01\\
	Tracking MPC & 0.99$\pm$0.03 & 0.75 $\pm$ 0.02\\
	Avoidance MPC &7.31$\pm$3.15&7.14 $\pm$3.01\\
	Our MPC&7.90$\pm$3.63&7.67$\pm$3.62\\
	\bottomrule
\end{tabular}
\vspace{1ex}
\caption{Comparison of multiple MPC formulations with respect to their update time and solver time.}
\vspace{-2.5ex}

\label{tab:mpc_run_time}
\end{table}
\textbf{Real-world deployment:}
We deploy our system on a quadrotor equipped with an Intel Realsense T265 tracking camera and an Intel Realsense D435i depth camera.
The onboard computer is an Nvidia Jetson TX2.
Detailed information about our quadrotor platform is in~\cite{foehn2022agilicious}.
We use the VIO algorithm from the tracking camera to obtain an estimate of the 6-DoF pose of the quadrotor and the depth camera to obtain RGB images for the line detection algorithm and depth measurements for the collision avoidance algorithm.
All the components of our system run on the onboard computer in real-time.
We set up a mock-up power line environment featuring 3 power masts.
The distance between each pair of power masts is 7.5 m.
A top view of the power environment including a trajectory flown by our quadrotor is in Fig~\ref{fig:catcheye}.
We run several experiments with different starting positions sampled in the vicinity of mast A.
\iros{In all the experiments, the proposed MPC does not have access to the ground-truth location of the power lines and power masts.
For this reason, a classical MPC approach without perception awareness is not a suitable solution.}
We demonstrate that our quadrotor is able to track the power line by adapting its heading (c.f., Fig~\ref{fig:catcheye}) and to avoid the power masts.
We refer the reader to the accompanying video for the visualization of these experiments.

\section{Conclusions} \label{sec:Conclusions}

In this work, we present a system for autonomous power line inspection using perception-aware MPC. 
Our approach generates control commands that maximize the visibility of the power lines while safely avoiding the power masts.
Our MPC formulation includes two perception objectives, one for line tracking and one for obstacle avoidance. 
The MPC adapts the weights of these two objectives online.
\iros{To detect the power lines, we propose a novel learning-based detector.
This learning-based detector is only trained on synthetic data and is able to transfer to real-world images without any fine-tuning.}
We show that our system is robust to unknown information on the position of the power lines and power masts.
We also show that our perception-aware MPC improves power line visibility by 40\%.
We demonstrate a real-world application in a mock-up power line environment.
Future improvements for our system might include i) improving system robustness against disturbances such as wind~\cite{cioffi2023hdvio}, model mismatch~\cite{bauersfeld2021neurobem} or sensor failures~\cite{sun2022see} and ii) perching on the power
line~\cite{paneque2022perception} in order to recharge the battery on the 
fly~\cite{kitchen2020design}.

{\small
\bibliographystyle{IEEEtran}
\bibliography{all}
}

\end{document}